\documentclass[10pt,twocolumn,letterpaper]{article}

\usepackage[pagenumbers]{cvpr} 

\usepackage{graphicx}
\usepackage{amsmath}
\usepackage{amssymb}
\usepackage{booktabs}

\usepackage[pagebackref,breaklinks,colorlinks]{hyperref}

\usepackage{paralist}

\DeclareMathOperator*{\E}{\mathbb{E}}

\usepackage[capitalize]{cleveref}
\crefname{section}{Sec.}{Secs.}
\Crefname{section}{Section}{Sections}
\Crefname{table}{Table}{Tables}
\crefname{table}{Tab.}{Tabs.}

\def\cvprPaperID{*****} 
\def\confName{CVPR}
\def\confYear{2022}

\begin{document}

\title{Data Distribution Shifts in (Industrial) Federated Learning as a Privacy Issue}

\author{David Brunner and Alessio Montuoro\\
Software Competence Center Hagenberg\\
Softwarepark 32a, 4232 Hagenberg im Mühlkreis, Austria\\
{\tt\small david.brunner@scch.at}, {\tt\small alessio.montuoro@scch.at}
}

\maketitle

\begin{abstract}
    We consider industrial federated learning, a collaboration between a small number of powerful, potentially competing industrial players, mediated by a third party aspiring to improve the service it provides to its customers. We argue that this configuration harbours covert privacy risks that do not arise in e.g. cross-device settings. Companies are very protective of their intellectual property and production processes. Information about changes to their production and the timing of which is to be kept private. We study a scenario in which one of the collaborators infers changes to their competitors' production by detecting potentially subtle temporal data distribution shifts. \textbf{In this framing, a data distribution shift is always problematic, even if it has no negative effect on training convergence.} Thus, our goal is to find means that allow the detection of distributional shifts better than customary evaluation metrics. Based on the assumption that even minor shifts translate into the collaboratively learned machine learning model, the attacker tracks the shared models' internal state with a selection of metrics from literature in order to pick up on relevant changes. In an empirical study on benchmark datasets, we show an honest-but-curious attacker to be capable of detecting subtle distributional shifts on other clients, in some cases long before they become obvious in evaluation.
\end{abstract}

\section{Introduction}\label{sec1}

\begin{figure}
\centering
\includegraphics[width=\linewidth,keepaspectratio]{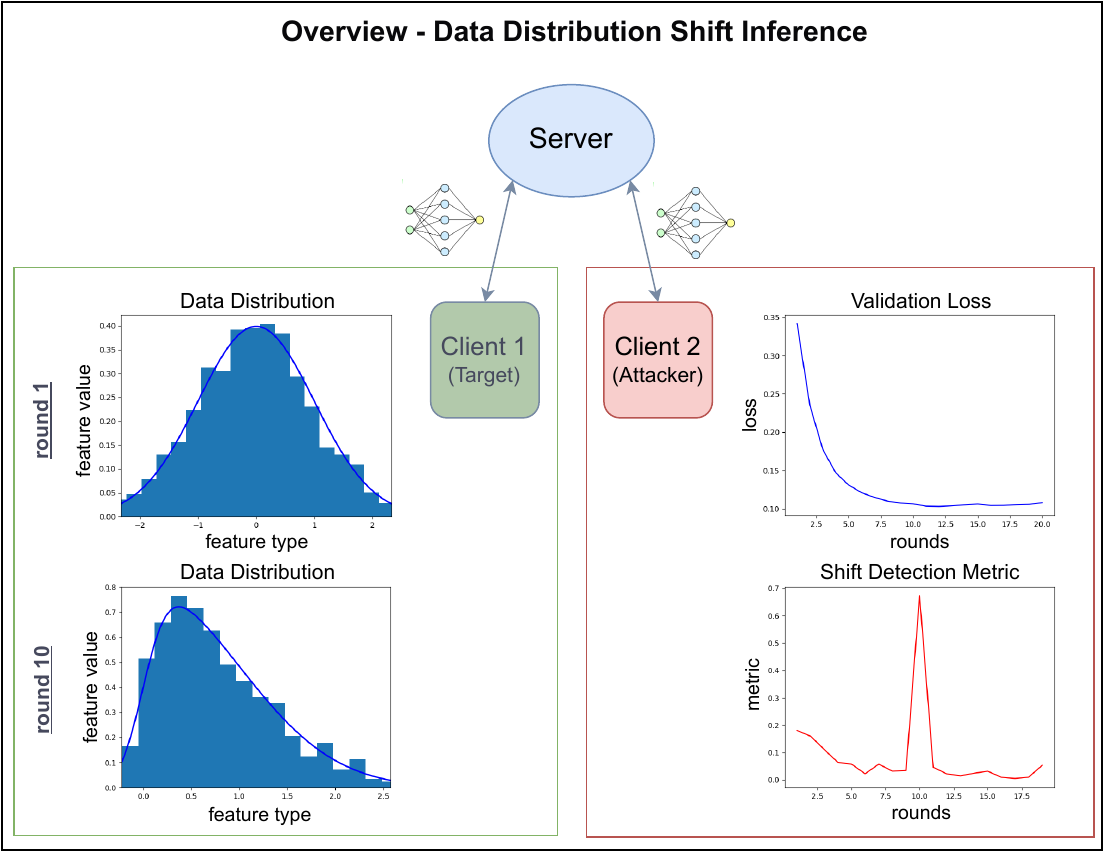}
\caption{High-level overview over the proposed attack on industrial FL. An attacker infers a data distribution shift on the other client occurring at some point during the FL. The shift is subtle enough so as to not affect conventional evaluation metrics, but can still be detected via more specialized metrics.}
\label{fig:overview}
\end{figure}

Since Federated Learning (FL) was introduced in 2017 by McMahan et al. \cite{mcmahan2017communication}, it has been studied extensively. In FL a model is trained on a number of data sources that are kept separate to preserve the privacy of their holders. This is done by instead of merging the data in a central location, only combining the models trained on each of the data sources respectively. The central location (the server) is responsible for aggregating the models into a joint, global model and returning it to the clients. After a sufficient number of rounds the resulting global model should behave similarly to a centralized model trained on the combined data sources. Much research has been dedicated to pressure testing FL's privacy guarantees, yielding an extensive list of attacks aimed at extracting sensitive information about the data of a client \cite{rodriguez2023survey, liu2022threats}. What comprises sensitive information depends on the context. Most works on FL focus on what is often referred to as \textit{cross-device} federated learning, best conceptualized as a business-to-consumer setting, in which e.g. an app developer improves their product by indirectly using the users' data. Comparatively little attention has been given to \textit{cross-silo} FL, which maps onto the business-to-business setting of collaborating companies \cite{huang2022cross}. While in both settings preserving the privacy of the clients is crucial, they differ in the fact that in the cross-silo setting, if two clients are eligible for collaboration (i.e. have similar data distributions), they most likely are competitors and therefore have especially high privacy demands. In other words, information leaks that are inconsequential in cross-device settings may be very consequential in cross-silo settings.

We identify data distribution shifts (DDS) as such a non-obvious privacy risk in FL applied to industrial collaborations. DDS are a well-known problem in FL and are typically viewed through the lense of their detrimental effects on training convergence. Crucially, the shifts discussed in this work are of temporal nature, i.e. occur at some point during the training process. Temporal shifts can occur when clients alter their data generating process. In industrial FL the clients are companies and a change in the data generating process could mean a change in production, like the introduction of a new material or the manufacturing of a new product. It is easy to see why this information would be considered sensitive\footnote{We acknowledge that the term \textit{intellectual property protection} would be more apt than \textit{privacy} for the information discussed in this work, since the affected entities are companies rather than individuals. However, to put this work in line with the terminology of existing attacks we opt to stick to privacy.}. Dramatic changes in a clients' data distribution will likely result in deteriorating training convergence, making them easily detectable by tracking the models' performance measures throughout the training. We show, however, that even changes subtle enough to not affect the training progress oftentimes can be inferred by another client by simple means and therefore pose potential privacy risks. Figure \ref{fig:overview} shows a high-level overview over the proposed threat model, which we lay out in detail in the next chapter. \\
In summary our contributions are the following:
\begin{itemize}
    \item We put forward a threat model for industrial FL, an understudied area of FL, in which a client can potentially gain sensitive information about their competitors manufacturing process.
    \item We introduce a simple inference attack on FL that aims at inferring temporal DDS on other clients.
    \item We perform experiments on benchmark datasets, showing that a subtle DDS can be inferred by a honest-but-curious attacker even if it does not have any impact on training convergence.
\end{itemize}

\section{Threat Model}\label{sec2}
\subsection{Industrial FL}
An early, successful, large-scale application of FL in practice is Google's GBoard \cite{bonawitz2019towards, yang2018applied}. Aiming at improving their products' next word prediction, Google had to handle millions of clients of which each only produces few datapoints, has limited computational power and drops in and out of the FL as the user loses their internet connection or turns off their phone. Now often referred to as \textit{cross-device} FL, this setting can be distinguished from the comparatively newer \textit{cross-silo} FL. In cross-silo FL the clients are few, but possess abundant data, high computational ability and are unlikely to drop out of the FL. Cross-silo FL often comes in the form of two hospitals deciding to collaborate without exposing their respective patients' data to each other. While collaborating hospitals provide a very intuitive example for cross-silo FL, it lacks one crucial feature of the precise setting we focus on in this work. In what we will refer to as \textit{industrial} FL, the clients are not only powerful institutions, but they are also potentially \textit{competitors}. Table \ref{table:FL-types} provides an overview of these variants. Consider a company $S$ whose business model it is to sell AI-powered predictive devices to their customers. The customers $C_{1}$ and $C_{2}$ are big manufacturers and use the devices to monitor their production. $S$ suggests to $C_{1}$ that by entering into a privacy-preserving collaboration with other manufacturers they can significantly improve their predictive accuracy. $C_{1}$ will only be convinced to enter the scheme if the highest level of privacy protection can be ensured. Even more so than the privacy of individual datapoints, the protection of global statistics and trends in their data will be of concern to $C_{1}$, the leaking of which could give $C_{2}$ a competitive edge, for instance by timing the manufacturing of their new product based on when $C_{1}$ appears to do so.

\begin{table}
\centering
\caption{\textbf{Our definition of industrial FL.} It represents a subtype of cross-silo FL in which the collaborators are potential competitors. *stability refers to the probability that a client participates in a round.}
\resizebox{\linewidth}{!}{%
\begin{tabular}{ l l l l }
\toprule
 property & cross-device & cross-silo & industrial FL \\ 
\midrule
num. clients & high & low & low \\  
num. data samples & low & high & high\\
 processing capabilities & low & high & high \\
stability* & low & high & high \\
competing clients & no & no & yes \\
\bottomrule
\end{tabular}
}

\label{table:FL-types}
\end{table}

\subsection{Data Distribution Shifts}
A core assumption we make is that the clients' data changes periodically over the course of the FL. As the data generating processes keep on providing new data, at some point in the FL, after training a reasonable number of rounds on the last chunk of data, it will be swapped with a new chunk. In absence of drastic changes in the data generating process the new chunk of data should follow the same statistical distribution as the previous ones. However, if there is a change in the data generating process the data distribution might also change. Since severe distributional shifts are exceedingly apparent in the training process, we are especially interested in minor shifts that would fly under the radar if only the conventional performance metrics (accuracy, loss) were to be consulted. In this vein, we also exclude concept drifts from our analysis. Formally, a dataset $D(X, Y)$ consists of features $X$ and labels $Y$. In a DDS either the feature distribution $P(X)$ or the label distribution $P(Y)$ changes while the conditional probability $P(Y|X)$ remains the same. Conversely, in a concept drift $P(Y|X)$ changes. In FL, if $P(Y|X)$ differs among the clients no joint model can be learned, leaving each client with poor performance and making concept drift easy to detect.

\begin{figure*}
\centering
\includegraphics[width=\textwidth,keepaspectratio]{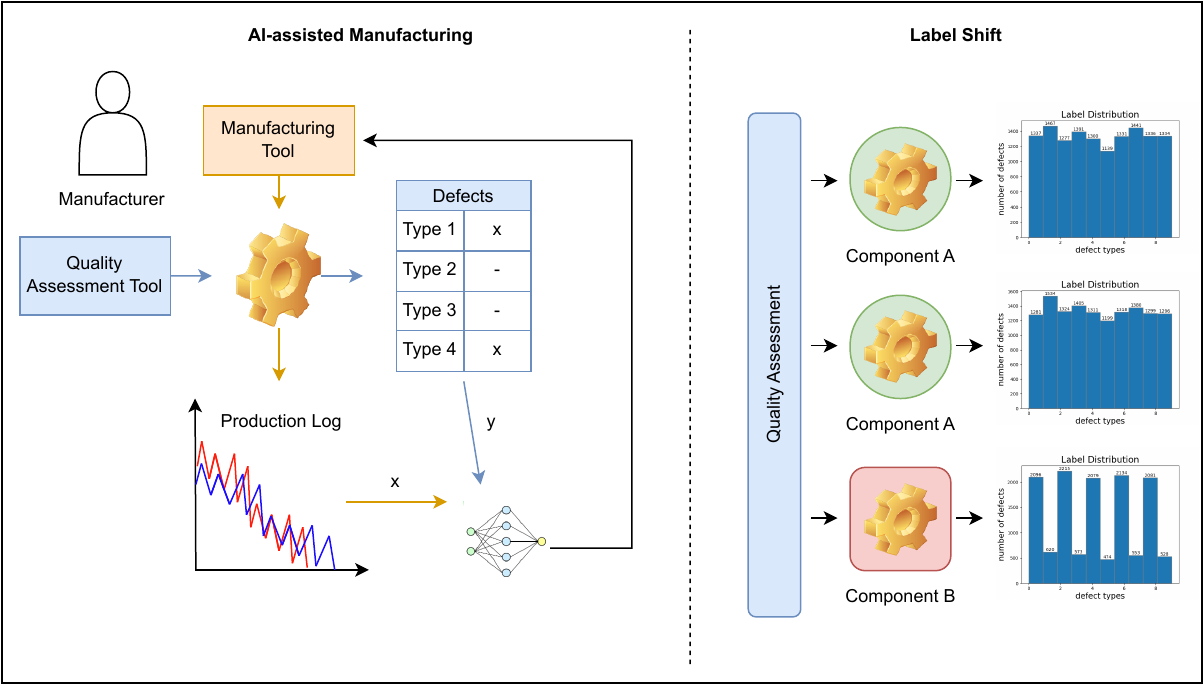}
\caption{\textbf{Left}: A company sells devices that assist the manufacturing of industrial components and allow the assessment of their quality. The company adds AI-based predictive services which allows the manufacturers to predict defects ahead of time by using accrued process- and quality data for training a model. Since the company's customers have similar tasks they could offer to join a collaborative scheme across manufacturers to enhance the predictive performance. \textbf{Right}: A label distribution shift can occur if a manufacturer produces a new component which affects the distribution of identified defects. The detection of these changes by other clients could pose a privacy risk by disclosing operational adjustments.}
\label{fig:dist-shift}
\end{figure*}

For an example of a realistic scenario in which DDS can occur, consider the following. A company sells devices used in the production of industrial components. The production of these components is intricate and can leave the components with defects that make them unusable. Defects are found after the fact via a separate quality assessment system, also provided by the company. As discarding components is costly, the manufacturers - the company's customers - would benefit greatly from a model that can predict ahead of time whether a given component will turn out flawed, so the process can be halted, saving time and raw material. In order to train such a model, for each produced component the manufacturers log process information and the accompanying quality assessment. Every so often the accrued data is passed to the customer's local model, the process information being the features and the quality assessment the labels. The company collects the trained models, aggregates and re-distributes them. As long as no dramatic changes are made to the production process the data distribution stays approximately the same. If, however, a customer starts producing a new component, initially, as they still lack the experience of how to optimally work the component, more defects might occur. Alternatively, due to a small innovation in the production process the customer might manage to slightly diminish the likelihood of defects. Both scenarios shift the label distribution and both scenarios describe changes, the information to which the customer (initially) may want to withhold. Figure \ref{fig:dist-shift} shows a visualization of the described scenario. The question now is, whether this change is noticeable to other clients in the FL, despite not showing up in their default performance measures. \textit{Is a subtle distributional shift on one client perceivable on other clients?}

\subsection{Modeling the attacker}
Existing attacks on FL differ in who the malicious actors are. Some attacks assume a malicious server, while in others a client takes the role of the attacker. In the industrial FL setting described above, it is the clients who have a vested interest in extracting private information from each other. We assume a malicious client to be honest-but-curious, trying to extract information while diligently following the protocol. In other words, the attacker is passive rather than active and manipulates neither data nor updates. The attacker's goal is to infer if at some point in the FL the other client's (the target's) data distribution has shifted. Figure \ref{fig:overview} depicts the general procedure. Both clients train on their local data and send updates to the server. At some point the target client swaps the local dataset. Since the new data follows the same distribution as the previous data, swapping should have no conspicuous effects on local and global model. After some updates to the data generating process, the target client swaps the local dataset again, this time with a shifted label distribution. Assuming the target client did not also update the local validation data along with the training data there are two possibilities for what happens next. Either 
\begin{inparaenum}[(i)]
\item the shift is severe enough to show up in the target client's local validation, allowing the client to retain the update in fear of information leakage (or simply negative effects on convergence), or
\item the shift is too weak to show up in the local validation and the target client proceeds with sending the update to the server as usual.
\end{inparaenum}
If the target client creates a new validation set from the newly arrived data, the shift will not show up in the validation anyway. If the shift is not apparent in the target client's validation, it is very unlikely that it is in the attacker's validation. To detect the shift, the attacker needs to find other means. In the next chapter we review the literature in search of existing techniques and subsequently detail our approach in chapter 4.

\section{Related Work}\label{sec3}

\subsection{Distribution Shift Detection} DDS are a prominent issue in machine learning. In centralized (i.e. non-FL) machine learning, shifts typically comprise the distribution of the test data diverging from the distribution of the training data. Since in this work we exclusively consider neural networks (NN) as model type, we exclude works on other model types from our literature review. In 2018 Lipton et al. \cite{lipton2018detecting} proposed Black Box Shift Estimation (BBSE) as a technique for detecting label shifts. BBSE is based on the insight that a shift in the input data distribution of a model will cause the model's output to shift as well. Specifically, they apply two-sample tests like Kolmogorov-Smirnov (KS) and Maximum Mean Discrepancy (MMD) to the confidence vector (i.e. output vector) distributions at training and test time, showing their divergence in case of label shifts. Viewed differently, instead of comparing the training and test data directly, the model serves as a means for dimensionality reduction, after which a shift can be detected more effectively. Based on this idea, Rabanser et al. \cite{rabanser2019failing} conducted a comprehensive study of two-sample tests in conjunction with various dimensionality reduction techniques, concluding that, overall, BBSE is indeed among the most effective. Bar-Shalom et al. \cite{bar2022distribution} tackle DDS detection by building on the idea of selective classification \cite{geifman2017selective}, in which a model can refuse giving a prediction if it is not confident enough. Their observation is, that a shift in the data distribution should lead to more uncertainty and hence cause a model to refuse to predict more often (i.e. exhibit less \textit{coverage}). They incorporate this notion into their method by first deriving a lower bound on coverage on data representative for the training distribution, and subsequently checking the violation of this bound on incoming windows of a stream of test data. Aside from improving on previous methods in terms of detection accuracy, they argue that the window based approach makes their method applicable in practice, while all previous methods' reliance on performing computations on the whole test set makes them computationally infeasible. Finally, Hensel et al. \cite{hensel2023magdiff} employ activation graphs \cite{lacombe2021topological} for DDS detection. They propose leveraging the much more expressive layer activations within a NN instead of the final confidence vectors as basis for two-sample tests. More precisely, they calculate what they call Mean Activation Graph Difference (MAGDiff), which allows quantifying the difference of the activations for a test sample to the mean activations across all training samples of a certain class.

\subsection{Out-of-Distribution Detection} Contrary to the DDS detection methods described above (which operate on a distributional level), Out-of-Distribution (OOD) detection tries to identify individual samples observed at test time as outside of the training distribution. This does not coincide exactly with the setting in this work, but some of the employed techniques are interesting nonetheless. In particular we want to mention \textit{ReAct} \cite{sun2021react} and \textit{GradNorm} \cite{huang2021importance}. The authors of ReAct (Rectified Activations) make the observation that OOD data leads to higher activation variation in the penultimate layer of a NN, and that an OOD score (they use the energy score) applied to these activations is more informative if they are truncated beforehand. GradNorm builds on the notion that OOD input causes the softmax output to be more uniform than In-Distribution (ID) input. By backpropagating from the Kullback-Leibler (KL) divergence between the softmax output and a uniform distribution gradients are obtained, which should be higher in magnitude in the case of ID data (since for ID data the softmax output does not resemble a uniform distribution).

\subsection{Poisoning Detection / Robust Aggregation} DDS detection in the context of FL can be viewed as a special case of poisoning detection. Poisoning in FL aims at either disrupting the training process or establishing certain unwanted behaviours in the victim model, which an attacker can achieve by manipulating the local dataset or the local model during the training phase. The DDS discussed in this work resembles inadvertent data poisoning, hence existing poisoning detection techniques seem relevant. However, all popular techniques \cite{shen2016auror, fung2018mitigating, zhao2020shielding, li2020learning, andreina2021baffle} task the \textit{server} with detecting poisoned updates, whereas in our setting detection has to take place on client-side. Furthermore, many of the above techniques use the validation loss as detection metric, while we are particularly interested in shifts that are not evident in the validation loss. A related class of methods is that of \textit{robust aggregation} in which anomalous updates are sought to be neutralized by statistical means. Techniques such as Krum \cite{blanchard2017machine}, or even more simple, median aggregation \cite{yin2018byzantine} are used. The downside of these techniques is slowed convergence and an unequal treatment of clients. Especially the latter may prove problematic in an industrial FL setting where comparable outcomes for all clients must be ensured (and might even be contractually stipulated).

\subsection{Clustered Federated Learning} Generally, in FL DDS are assumed to be spacial, i.e. that data distributions differ among the clients, and do so from the start, for the simple reason that the datasets are of different origin. In some cases the shift is severe enough that training a combined model among all clients makes no sense. A useful strategy for these cases is to train not one, but multiple global models, each shared by clients with similar data distributions. Sattler et al. \cite{sattler2020clustered} proposed a method called \textit{Clustered Federated Learning} (CFL), which groups clients into clusters based on the cosine similarity of their gradient updates. They explicitly state that CFL underperforms for subtle shifts, since it is aimed specifically at concept drifts. Furthermore, they do not take into account shifts occurring \textit{during} the FL. Jothimurugesan et al. \cite{jothimurugesan2023federated} fill this gap by developing a method for dealing with \textit{temporal} concept drifts. However, they again rely on the validation loss for estimating the differences of the clients data distributions, a metric which we expressedly rule out for our attack.

In the next chapter we develop our attacker's approach.

\section{Method}\label{sec4}
\subsection{Attacker Information}

\begin{figure}
\centering
\includegraphics[width=\linewidth,keepaspectratio]{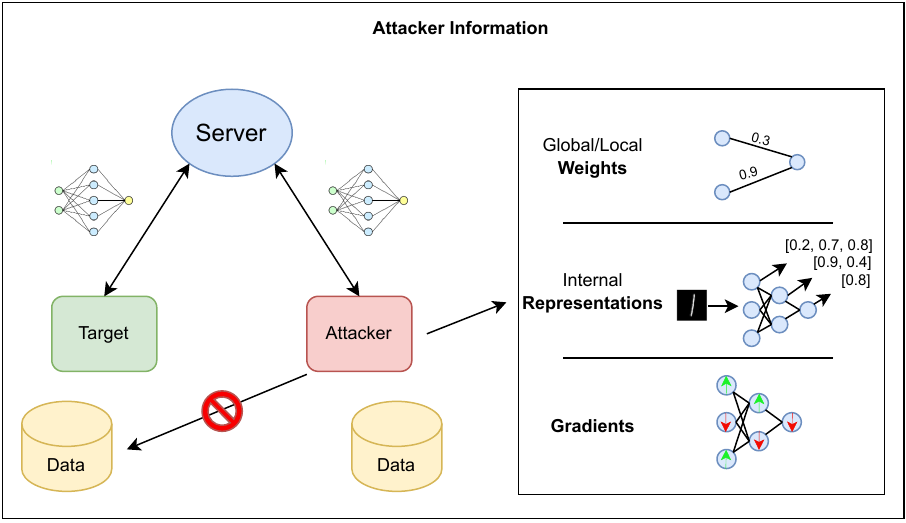}
\caption{\textbf{Available information for an attacker in FL.} The attacker cannot access the data of another client directly, but the global model's internal state is accessible, including the weights, the internal representations and the gradients. Ideally, these encode enough evidence of a DDS for the attacker to extract.}
\label{fig:attacker-info}
\end{figure}

The task is to find a method or metric, that allows the detection of DDS on other clients \textbf{more accurately} than the standard performance measures, namely validation loss and validation accuracy. Most of the existing DDS detection methods require access to a trained model as well as data from before and after the shift (i.e. train and test data). In FL, the shifted data is shielded from other clients and therefore inaccessible for the attacker. However, the relation of global model to local data still changes, only that from the attacker's perspective it is the model that (affected by the shift) changes while the data is static. In other words, instead of gauging the change in data based on a static model, the attacker tries to gauge the change in a model via static data. The latter is of course equivalent to validation, only that we are interested in metrics other than model performance, and even though the validation loss can reflect a DDS, the goal is to find more expressive metrics. While early DDS detection methods like BBSE inspect only the outputs of the last layers of a NN, we take the approach of scrutinizing the internals of the NN. The simplest possible approach is to analyze the weights directly, the idea being that the weights of the global model in subsequent rounds differ more clearly from one another when the training data experienced a shift, as opposed to when it did not. Another already discussed option is presented by Hensel et al. \cite{hensel2023magdiff}, who show the informativeness of the layer activations with respect to detecting DDS. The layer activations can be obtained by passing data to the NN and collecting the outputs of each layer, thus revealing how the data is internally represented by the NN. For this reason we will refer to them as internal \textit{representations} in the remainder of this work. Again, the idea is to compare the representations across subsequent rounds, by passing static data to a changing model. Finally, inspired by Sattler et al. \cite{sattler2020clustered}, we examine the gradients as a source for evidence of a DDS. The gradients are much more reactive to change than, say, the weights, which makes them attractive for the task at hand. We conclude that the types of \textit{attacker information} are threefold:
\begin{inparaenum}[(i)]
\item the global/local weights,
\item the internal representations and
\item the gradients, 
\end{inparaenum}
see Figure \ref{fig:attacker-info} for a visual summary.

\begin{figure}
\centering
\includegraphics[width=\linewidth]{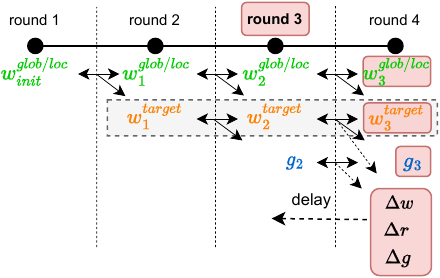}
\caption{\textbf{Timeline of attacker information acquisition.} In the second round the attacker can extract the target weights and obtain the representations, and approximate the gradients in the third. Their trend can then be determined one round later respectively, by comparing the values of subsequent rounds. Therefore, a DDS can be detected by means of weight- and representation trend from round two forward and via the gradients starting from round three. Since the global weights of any specific round always encode the contributions of the clients from the round before, the detection of a DDS always has a one round delay.}
\label{fig:info-acqu}
\end{figure}

The process of obtaining the attacker information then looks like the following. In the first round of the FL the attacker saves the global weights (i.e. before local training, which would diminish the target's influence), as well as the representations, obtained by querying the global model with a representative subset of the validation or test data\footnote{In a real world scenario the attacker does not have access to samples of the client's data distribution. However, since an industrial collaboration only makes sense if the collaborators run similar businesses and therefore process similar data, it is plausible that the attacker's data is sufficiently representative for an altered model to produce meaningfully altered representations.}.  In the second round, the current weights and representations can be compared to their predecessors of the round before. The gradients can be approximated by subtracting the weights of the current round from their predecessors. In order to then compare the gradients of subsequent rounds, another round has to be completed. Optionally, the attacker can also try to remove their own influence from the global model by reversing the aggregation operation of the server and thereby "extract" the target model from the global model. We consider the FedAvg \cite{mcmahan2017communication} protocol, for which aggregation simply consists in an averaging step. To remove their influence from the averaged model, the attacker must only know the number of clients participating in a round of FL, which is a reasonable assumption for industrial FL. Figure \ref{fig:info-acqu} sketches the acquisition timeline of the attacker information.

\subsection{Shift Metrics}
In this section we address the yet unanswered question of how to quantify the change in the attacker information across subsequent rounds. In the DDS detection literature metrics like MMD are leveraged to quantify the shift in the data distribution. In absence of access to the data, our attacker operates based on the assumption that the shift in the data's distribution translates more or less directly into the model. While it is not obvious that that would be the case, given the stochasticity of the learning process, existing literature suggests that this assumption is sound. Sattler et al. \cite{sattler2020clustered}, for instance, discovered that gradient updates of clients with similar data distributions look similar. This matches our situation perfectly. They use \textit{cosine similarity} (\ref{equ:cs}) to compare the updates, which gives our attacker the first metric:

\begin{equation}
  cosine\_similarity(A, B)=\dfrac {A \cdot B} {\left\| A\right\| \left\| B\right\|},
  \label{equ:cs}
\end{equation}
where $A$ and $B$ are attacker information vectors of two NNs.

More generally, dissimilarity metrics have been studied as a tool for analyzing NN behaviour, like the impact certain hyperparameters have on the internal representations. A metric that responds to changes in the representations caused by modified hyperparameters, should also be able to indicate changes caused by shifted training data. Ding et al. \cite{ding2021grounding} provide a comprehensive analysis of existing dissimilarity metrics and conclude that all popular metrics have important failure modes, while the \textit{procrustes distance} (\ref{equ:pd}), their experimental baseline, gives surprisingly stable results. Motivated by this insight, we include the procrustes distance in our evaluation:

\begin{equation}
  procrustes(A,B)=\operatorname{trace}\left(X^{T} X\right)^{\frac{1}{2}},
  \label{equ:pd}
\end{equation}
where $X=A-B$ and $A$ and $B$ are attacker information vectors of two NNs.

Finally, we want to include a moment based metric, however, instead of the classic MMD we use the more effective and computationally efficient Central Moment Discrepancy (CMD) \cite{zellinger2017central} (\ref{equ:cmd}):

\begin{equation}
  CMD(A,B)={\|\E(A)-\E(B)\|_2} + \sum_{k=2}^{K} \|C_k(A) - C_k(B)\|_2,
  \label{equ:cmd}
\end{equation}
where $A$ and $B$ are attacker information vectors of two NNs, $\E(A)=\frac{1}{|A|}\sum_{a \in A}a$ is the empirical expectation vector of $A$, and $C_k(A)=\E((a-\E(A))^k)$. We set $K=5$ in all our experiments.


We will refer to the combinations of these \textit{shift metrics} and attacker information as \textit{Sources of Leakage} (SoL). In the next chapter we report on performed experiments, evaluating the viability of the proposed approach. First, we construct a simplified setting with a centralized model in order to study the SoL in a targeted manner. Based on the findings of this first stage we then conduct experiments in the FL setting.

\section{Experiments}\label{sec5}
\subsection{Experimental Settings}
\paragraph{Platform} All experiments were performed on a single A100 GPU. The code was developed on Ubuntu 20.04 and tested on Ubuntu 18.04 and 20.04, using PyTorch\footnote{\url{https://pytorch.org}} 1.13 for training the models and Flower\footnote{\url{https://flower.dev/}} 1.3  as the FL framework.
\paragraph{Datasets} We conduct experiments with 3 datasets: MNIST\footnote{\url{http://yann.lecun.com/exdb/mnist}}, Fashion-MNIST \cite{xiao2017fashion} and Census\footnote{\url{https://www.kaggle.com/datasets/uciml/adult-census-income}}.
\paragraph{Architectures}  We use a 4- layer convolutional NN with a total of 1.2M parameters for MNIST and Fashion-MNIST. For Census we use a 5-layer fully-connected NN with 3.6K total parameters. All architectures use softmax activations in the final layer and ReLU in all others. The architectures are detailed in table \ref{table:arch}.

\begin{table}
\centering
\caption{\textbf{The NN architectures for each dataset.} The number in parentheses denotes the number of layer neurons. In the case of convolution layers the kernel size and stride are also provided.}
\begin{tabular}{ l l }
\toprule
Fashion-/MNIST & Census \\ 
\midrule
conv (32, 3, 1) & linear (64) \\  
conv (64, 3, 1) & linear (32) \\  
flatten & linear (16) \\  
linear (128) & linear (8) \\  
linear (10) & linear (2) \\ 
\bottomrule
\end{tabular}
\label{table:arch}
\end{table}

\begin{table}
\centering
\caption{\textbf{Experiment parameter overview.} Note that $n$, $m$ and $l$ only apply to the FL experiments.}
\begin{tabular}{ l l }
\toprule
parameter & meaning \\
\midrule
$r$ & epochs/rounds \\ 
$s$ & shift epoch/round \\
$d$ & dataset size \\ 
$n$ & number of clients \\
$m$ & number of shifting clients\\
$l$ & local epochs \\
\bottomrule
\end{tabular}
\label{table:params}
\end{table}

\subsection{Centralized Model}
\begin{figure}%
    \centering
    \subfloat[\centering The similarity of the types of attacker information of subsequent epochs as measured by the shift metrics (i.e. the SoL). All SoL show the DDS.]{{\includegraphics[width=\linewidth]{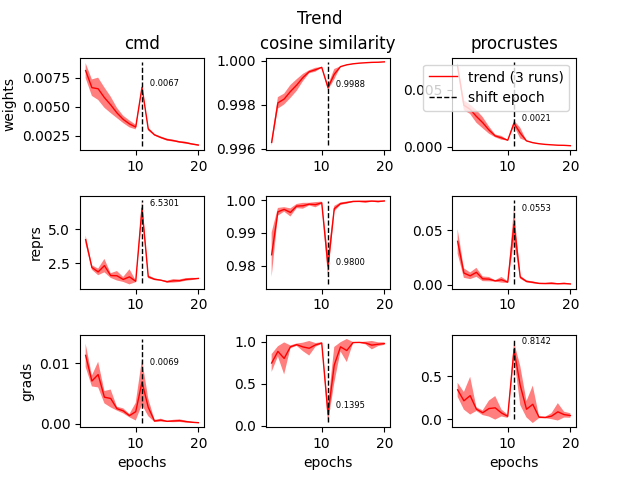} }}%
    \qquad
    \subfloat[\centering The validation loss does not betray the DDS.]{{\includegraphics[width=\linewidth]{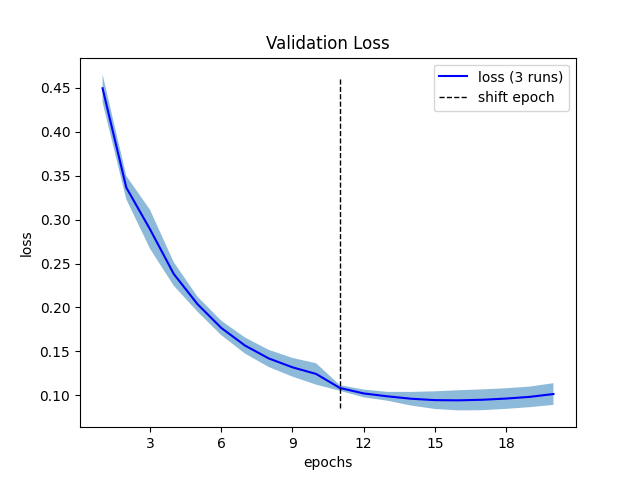} }}%
    \caption{\textbf{Results of the centralized setting.} The plots show a study of the SoL on a NN trained on MNIST, experiencing a label distribution shift at round 11. All results are reported as mean and standard deviation across three runs.}%
    \label{fig:central-results}%
\end{figure}

Before moving to the more complex FL setting we study the proposed method in a centralized setup. The effects of a DDS are experienced more directly in an isolated model, in absence of other models' influence. The experimental setup takes the following shape. A NN is trained on MNIST for $r$ epochs, during which validation and SoL are logged. At epoch $s$ the data, consisting of $d = 6.7K$ samples, is swapped with data of a shifted distribution. For our experiments we simulate a label distribution shift, by shifting the ratio of even to odd numbers in the training dataset. At epoch $s$ the original ratio (which is close to 50-50) shifts to 80-20. We set $r=20$ and $s=11$ in all experiments. We use the Adam optimizer with learning rate = 0.0001 and the crossentropy loss function. For an overview over the parameters refer to table \ref{table:params}. Figure \ref{fig:central-results} shows the results. Note that the shift metrics have varying scales. The procrustes distance adheres to the interval $[0, 1]$, cosine similarity to $[-1, 1]$ and CMD to $[0, \infty]$.

\begin{figure}
\centering
\includegraphics[width=\linewidth]{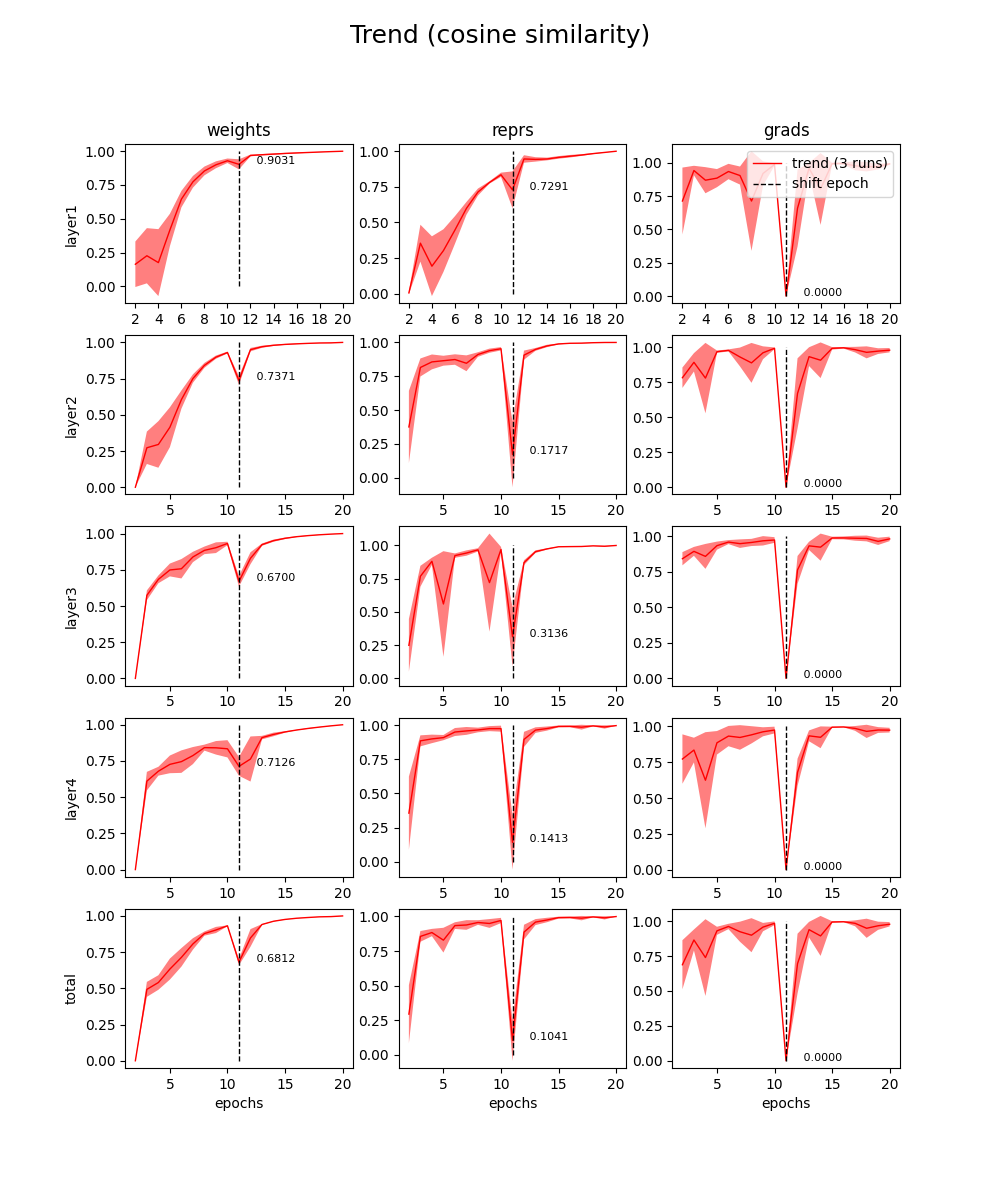}
\caption{\textbf{Layer-wise study of the cosine similarity metric in the centralized setting.} Weights and representations show less reaction to DDS in earlier layers. All results are normalized and reported as mean and standard deviation across three runs.}
\label{fig:layer-study}
\end{figure}

The results do not evince the clear superiority of any specific SoL, but indicate that the weights lack the sensitivity of the representations and gradients towards detecting a DDS. To answer the further question if individual layers are more expressive in terms of predicting a DDS we extend the previous experiment, evaluating the cosine similarity on a per-layer basis. The normalized per-layer metrics are shown in Figure \ref{fig:layer-study}. Generally, later layers prove to be more informative of a DDS than earlier layers. However, on the whole the results give no reason to prefer the measurements on individual layers over those computed across the full model.

In the next section, we study the FL setting, where we put the gained insights to use.

\subsection{FL Model}
The FL setting comprises $n$ clients - one of which takes the role of the attacker - training on their local dataset of size $d$ for $l$ epochs, and collaborating for $r$ rounds via the FedAvg protocol. Initially, the data is split IID across clients. At round $s$, $m$ clients (other than the attacker) experience a DDS. The attacker tracks the SoL and attempts to remove their own influence beforehand in all experiments. With this setup we conduct two types of experiments. First, we compare the sensitivity of the SoL to the sensitivity of the validation loss towards a DDS. Second, we investigate how well the inference scales with a growing number of clients. To achieve the former, we measure the divergence of the SoL and the validation loss from their respective trends at round $s$. We do this by linearly extrapolating the expected value at round $s$ from the preceding $e$ rounds and compare it to the measured value. If the SoL are more indicative of a DDS, then the divergence of the SoL from its trend should be more extreme than the divergence of the validation loss from the validation trend.

\begin{figure}%
    \centering
    \subfloat[\centering The sensitivity of the gradients as measured by the CMD on \textbf{MNIST}.]{{\includegraphics[width=\linewidth]{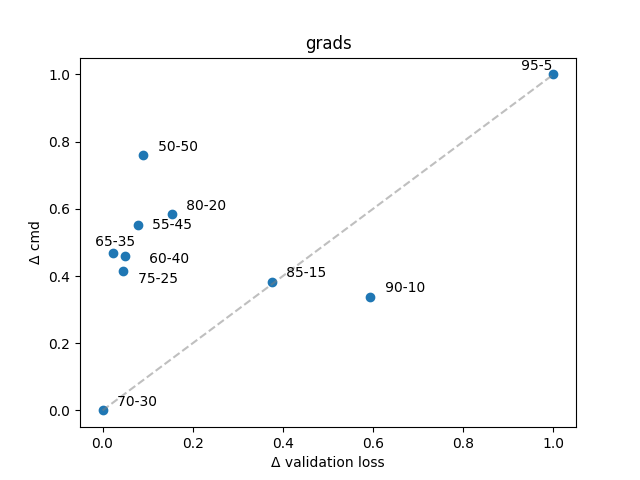} }}%
    \qquad
    \subfloat[\centering The sensitivity of the gradients as measured by the cosine similarity on \textbf{Fashion-MNIST}.]{{\includegraphics[width=\linewidth]{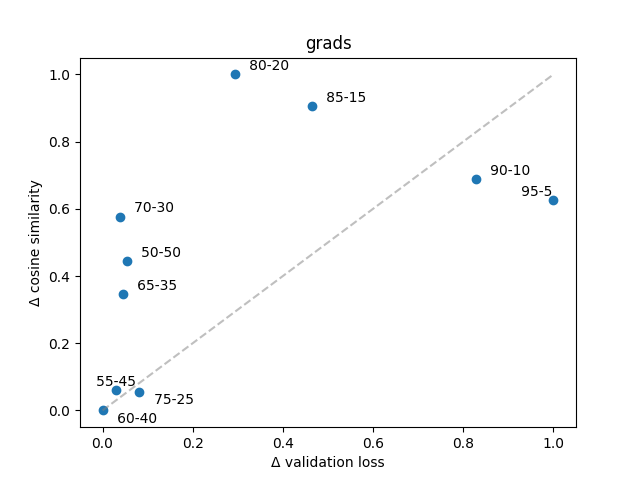} }}%
    \qquad
    \subfloat[\centering The sensitivity of the representations as measured by the CMD on \textbf{Census}.]{{\includegraphics[width=\linewidth]{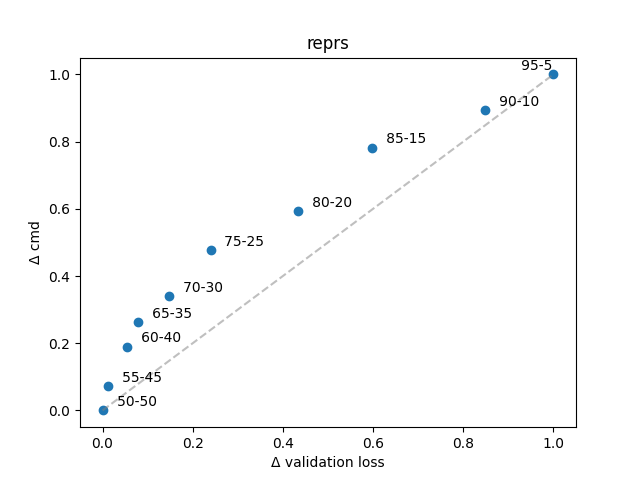} }}%
    \caption{\textbf{Sensitivity study.} Points falling above the diagonal line indicate the SoL being more sensitive than the loss. We are especially interested in the cases where the divergence of the validation loss is close to zero, but the divergence of the respective SoL is not. All results are reported as mean across three runs.}%
    \label{fig:fl-results-analysis}%
\end{figure}

\paragraph{Sensitivity Study} In the sensitivity study, the common settings across datasets are $n = 2$, $l = 2$, and $m = 1$. For MNIST, Fashion-MNIST and Census $r = [20, 40, 100]$, $s = [11, 21, 51]$, and $d = [6.7K, 6.7K, 3.9K]$ respectively. The shift type represents a label shift in all experiments, altering the ratio of even to uneven classes while keeping the number of samples constant. We repeat this process for increasing shift severities, the results can be seen in Figure \ref{fig:fl-results-analysis}. The type of SoL giving the best results varies across datasets. 

\begin{figure}%
    \centering
    \subfloat[\centering The trend of the gradients as measured by the cosine similarity.]{{\includegraphics[width=\linewidth]{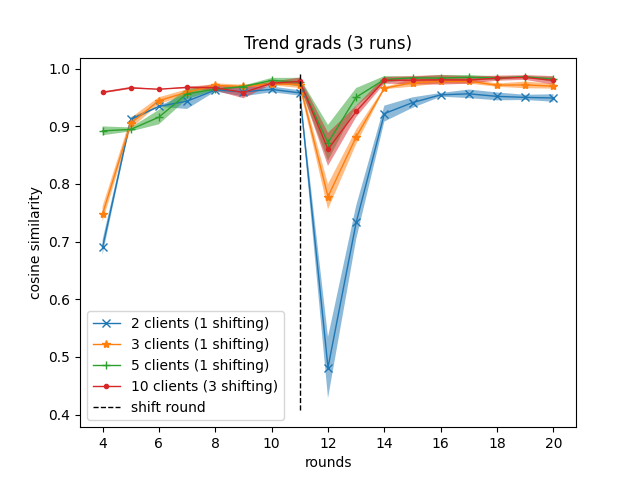} }}%
    \qquad
    \subfloat[\centering The validation loss does not show signs of a DDS.]{{\includegraphics[width=\linewidth]{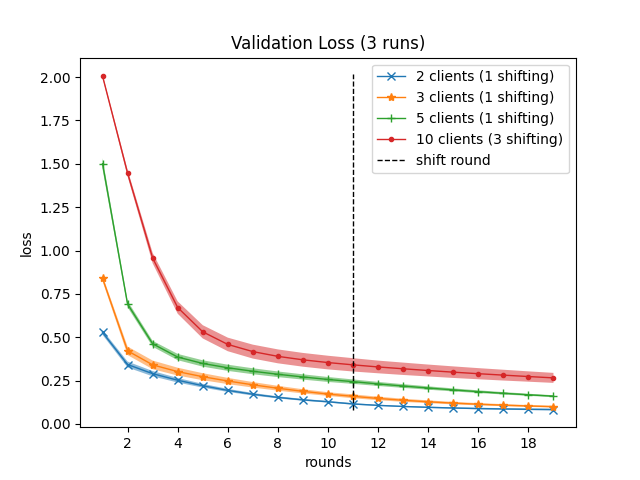} }}%
    \caption{\textbf{Scalability study on MNIST.} The plots show a study of the scaling properties of the proposed approach on a NN trained on MNIST, experiencing a label distribution shift at round 11. All results are reported as mean and standard deviation across three runs.}%
    \label{fig:mnist-num-cl}%
\end{figure}

\begin{figure}%
    \centering
    \subfloat[\centering The trend of the gradients as measured by the cosine similarity.]{{\includegraphics[width=\linewidth]{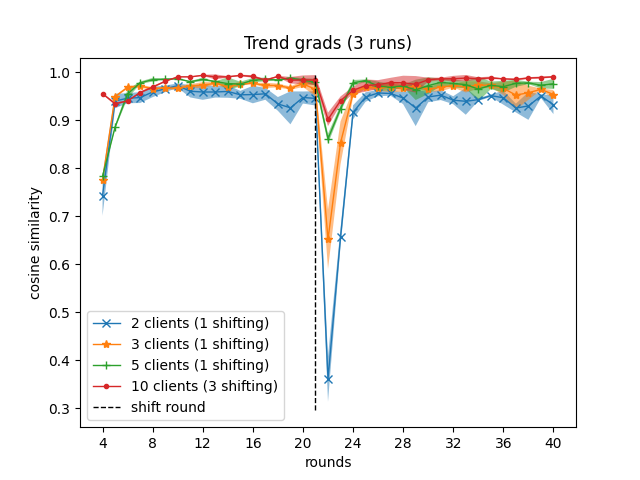} }}%
    \qquad
    \subfloat[\centering The validation loss does not show signs of a DDS.]{{\includegraphics[width=\linewidth]{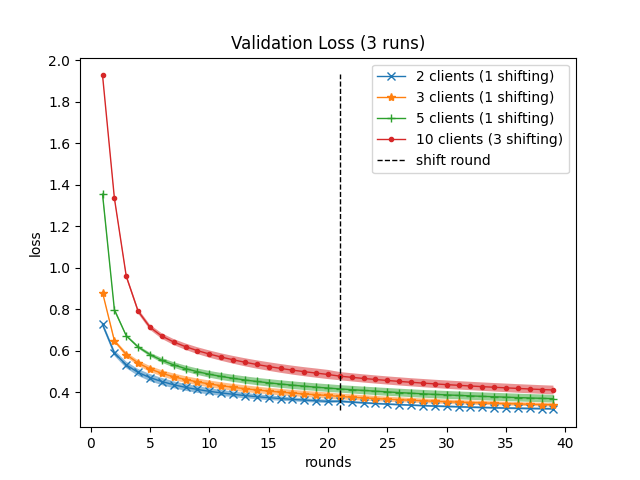} }}%
    \caption{\textbf{Scalability study on Fashion-MNIST.} The plots show a study of the scaling properties of the proposed approach on a NN trained on Fashion-MNIST, experiencing a label distribution shift at round 21. All results are reported as mean and standard deviation across three runs.}%
    \label{fig:fmnist-num-cl}%
\end{figure}

\begin{figure}%
    \centering
    \subfloat[\centering The trend of the representations as measured by the CMD.]{{\includegraphics[width=\linewidth]{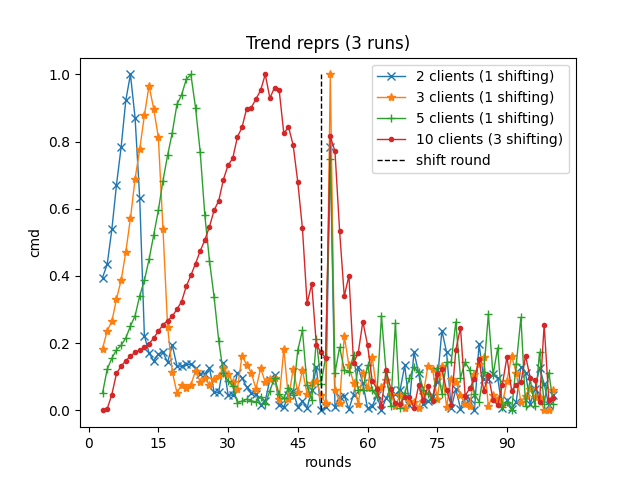} }}%
    \qquad
    \subfloat[\centering The validation loss does not show signs of a DDS.]{{\includegraphics[width=\linewidth]{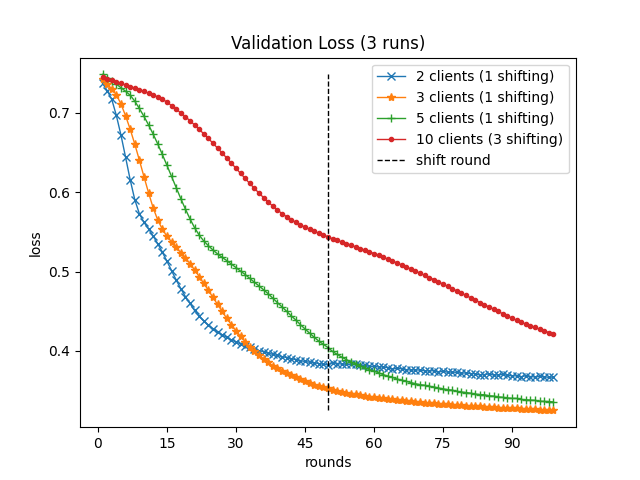} }}%
    \caption{\textbf{Scalability study on Census.} The plots show a study of the scaling properties of the proposed approach on a NN trained on Census, experiencing a label distribution shift at round 51. All results are reported as mean and standard deviation across three runs.}%
    \label{fig:census-num-cl}%
\end{figure}

\paragraph{Scalablity Study} In the scalability study, again $r = [20, 40, 100]$, $s = [11, 21, 51]$, and $l = 2$ for MNIST, Fashion-MNIST and Census respectively. We perform four experiments for each dataset, with $n = [2, 3, 5, 10]$, and $m = 3$ in the case of ten clients and $m = 1$ in all other cases. For MNIST and Fashion-MNIST $d = [6.7K, 4.4K, 2.6K, 1.3K]$, and for Census $d = [3.9K, 2.6K, 1.5K, 0.8K]$. At round $s$ the ratio of even to odd numbers shifts to 70-30 in the case of MNIST and Fashion-MNIST, and to 60-40 in the case of Census. Note that the shifted ratio is relative to the original ratio, which in the case of Census is already unbalanced (i.e. 60-40 does not mean the labels are actually split exactly 60-40). The results of the experiments can be seen in Figures \ref{fig:mnist-num-cl}, \ref{fig:fmnist-num-cl} and \ref{fig:census-num-cl}. We discuss the results of both studies in detail in the next chapter.

\section{Discussion}\label{sec6}
\subsection{Unveiling subtle distributional shifts}
The usefulness of the proposed attack hinges on the answer to the question, whether there are cases in which a DDS remains undetected with the conventional evaluation metrics, but can be inferred by more specialized means. The experimental results show this to be the case. In the sensitivity study on MNIST and Fashion-MNIST (Figure \ref{fig:fl-results-analysis}a and \ref{fig:fl-results-analysis}b), only severe shifts noticeably impact the validation loss, while even rather subtle shifts affect the trend of the gradients as measured by the CMD and the cosine similarity respectively. We also observe, that empirically the SoL do not reflect the severity of the shifts accurately, with subtle shifts scoring higher than more severe shifts at times. Conversely, the results on Census show the representations as measured by the CMD to have a linear relationship between shift severity and divergence from the trend. However, while the SoL consistently outperforms the validation loss, the latter does not stay completely oblivious to subtle shifts as is the case for MNIST and Fashion-MNIST, which suggests that the potency of the attack varies across datasets. 

In FL, changes introduced by one client inherently get diluted by the contributions of the other clients. For this reason, a large number of clients presents a natural obstacle for our attack. The results of the scalability study in Figure \ref{fig:mnist-num-cl}, \ref{fig:fmnist-num-cl} and \ref{fig:census-num-cl} show that detection works reasonably well for client counts that are realistic for industrial FL. DDS are clearly discernible even in the most challenging setting, but it should also be clear, that the attack does not scale up to thousands of clients. The results on Census also show a peculiar behaviour, as the model appears to take a number of rounds before finding a stable internal representation, only to be challenged by the DDS again. Such behaviour can be distinguished from a DDS by the more gradual nature of the change, contrary to the abrupt spike caused by the DDS.

\subsection{Limitations and Future Work}
The proposed attack is confined to a very specific setting. Only when there is demand for very high privacy standards does the inferred information present a threat. The experimental results also suggest the attack does not scale well to a large number of clients. While these assumptions seem limiting, they exactly match the scenario the attack was designed for: industrial FL. In industrial FL the number of clients is low by definition and privacy of utmost concern. In other words, though the attack is not applicable to a large variety of scenarios, it has great impact in those it is. Beyond that, it comes with convenient properties. For instance, no server involvement is presupposed, the attack entirely transpires on the attacker client. Also, the attacker does not have to have access to any kind of information apart from what is involved in a standard FL process.

Still, unexplored areas remain, which we leave for future work. For one, it is interesting to consider how well the attack scales to larger, more complex architectures. Another lane of research is to design an active attacker, who can launch more powerful attacks by manipulating the FL protocol. Finally, we excluded privacy-preserving techniques from this work. While cryptographic techniques like \textit{homomorphic encryption}, aimed at protecting client privacy from a curious server are not applicable as defence against a client-side attack, noise-adding techniques like \textit{differential privacy} seem very relevant.

\subsection{Practical Impact}
Federated Learning as a scientific field is a soaring, as attested by the overwhelming number of publications in the last few years. However, the practical adoption of the proposed techniques lags behind, which is especially true for industrial FL. From a company's perspective, an incentive for a participation in such an FL scheme can only arise if certain criteria are met. Only if the partner companies' processes and data are of sufficient similarity there is a prospect of noteworthy gains. Sharing a common trade, however, means that the participating companies already have a naturally good understanding of each others businesses and seemingly inconsequential leakage of information can result in a competitive advantage. The attack proposed in this work threatens one such type of information: the timing of manufacturing processes. Depending on the kind of business, the idea of a competitor being able to make an educated guess about the timing of one's own upcoming new product lines way in advance, might be intolerable. From the inquiring company's perspective it might not even be necessary to pinpoint a specific competitor's shift, only to pick up on a subtle, sudden shift in the population. We argue, that in industrial FL the stakes are high enough for concerns like these to deter companies from committing to a collaboration. In order for industrial FL to find widespread adoption it will be necessary to make threats of even this kind be implausible beyond reasonable doubt. We hope that this paper inspires further work on hidden threats in industrial FL.

\section{Conclusion}\label{sec7}

We showed that in industrial FL - cross-silo FL with competing parties - an honest-but-curious attacker can extract sensitive information about other clients data generating processes by inferring temporal data distribution shifts based solely on information native to FL. The performed experiments indicate that while the attack lacks scaling abilities, it performs well within the boundaries of the discussed setting and allows the detection of data distribution shifts that would go undetected otherwise.

\paragraph{Acknowledgement}
The research reported in this paper has been supported by the Austrian Research Promotion Agency (FFG), FFG Grant PRIMAL (Privacy Preserving Machine Learning for Industrial Applications) and by BMK, BMAW, and the State of Upper Austria in the frame of the SCCH competence center INTEGRATE [(FFG grant no. 892418)] part of the FFG COMET Competence Centers for Excellent Technologies Programme.

{\small
\bibliographystyle{ieee_fullname}
\bibliography{egbib}
}

\end{document}